\documentclass[letterpaper, 10 pt, conference]{ieeeconf}  % Comment this line out if you need a4paper

\IEEEoverridecommandlockouts                              % This command is only needed if 
\usepackage{inputenc}
\usepackage{hyperref}
\usepackage{amsmath}
\usepackage{graphicx}

\usepackage{afterpage}
\usepackage{color}
\usepackage{xcolor,colortbl}

\usepackage{pdflscape}

\title{\LARGE \bf
A Whole-Body Pose Taxonomy for Loco-Manipulation Tasks}

\author{J\'ulia~Borr\`as~and~Tamim~Asfour     % <-this % stops a space
\thanks{The authors are  with the Institute for Anthropomatics and Robotics, Karlsruhe Institute of Technology, Germany,
        {\tt\small \{julia.borrassol, asfour\}@kit.edu}}%
}

\begin{document}

\maketitle
\thispagestyle{empty}
\pagestyle{empty}

%%%%%%%%%%%%%%%%%%%%%%%%%%%%%%%%%%%%%%%%%%%%%%%%%%%%%%%%%%%%%%%%%%%%%%%%%%%%%%%%
\begin{abstract}

Exploiting interaction with the environment is a promising and powerful way to enhance stability of humanoid robots and robustness while executing locomotion and manipulation tasks. Recently some works have started to show advances in this direction considering humanoid locomotion with multi-contacts, but to be able to fully develop such abilities in a more autonomous way, we need to first understand and classify the variety of possible poses a humanoid robot can achieve to balance. To this end, we propose the adaptation of a successful idea widely used in the field of robot grasping to the field of humanoid balance with multi-contacts: a whole-body pose taxonomy classifying the set of whole-body robot configurations that use the environment to enhance stability. We have revised criteria of classification used to develop grasping taxonomies, focusing on structuring and simplifying the large number of possible poses the human body can adopt. We propose a taxonomy with 46 poses, containing three main categories, considering number and type of supports as well as possible transitions between poses. 

The taxonomy induces a classification of motion primitives based on the pose used for support, and a set of rules to store and generate new motions. We present preliminary results that apply known segmentation techniques to motion data from the KIT whole-body motion database. Using motion capture data with multi-contacts, we can identify support poses providing a segmentation that can distinguish between locomotion and manipulation parts of an action.

\end{abstract}

%%%%%%%%%%%%%%%%%%%%%%%%%%%%%%%%%%%%%%%%%%%%%%%%%%%%%%%%%%%%%%%%%%%%%%%%%%%%%%%%
\section{INTRODUCTION}\label{sec:Introduction}
%Humanoids motion planning, decision making, perception... parallelism to grasping
In the last decades, we have seen many successful results to make humanoid robots walk, stand and run maintaining balance~\cite{kajita_biped_2010, pratt_capture_2006, wieber_stability_2002}. However, such movements are still not robust enough for real life environments.
% and usually the robot avoids contact with the environment to prevent falling 
While  contacts with the arms can be used to achieve more stable poses and movements, the problem of generating whole-body motions with multi-contacts is very challenging and the traditional ZMP approach cannot be used. Several approaches exist, either using complex planning algorithms with constraints or using control methodologies to handle multi-contact situations with impressive results \cite{ 
sentis_2010_compliant_TRO, saab_dynamic_2013, lengagne_2013_generation, Hirukawa_2006_AdiosZMP}. However, these solutions are still computationally very expensive  and not able to perform autonomous locomotion tasks with multiple contacts. 

\begin{figure}[t!]
\includegraphics[width=\columnwidth]{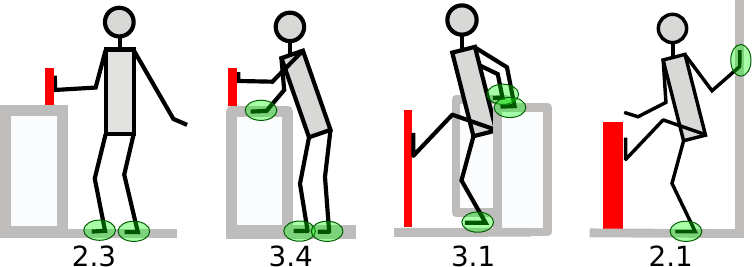}%
\caption{To perform the task of hiting the red object, several poses can be chosen to provide the stability that the task requires. The support poses are defined by the contacts highlighted in green. The numbers under the sketches refer to the id number of the support class according to our taxonomy in Table \ref{fig:taxonomy}}
\label{fig:HitWithLeg}%
\end{figure}

Recent results in simulation of walking in unstructured environments show a humanoid using walking staff~\cite{khatib_suprapeds_2014}. Using the staff allows to use combinations of two, three and four support poses to obtain a more stable walking. However, the robot could gain stability in many different ways, not only using tools but also its own body. Understanding all the possible ways the body can be used for balancing can be crucial not only to provide more stable motions, but to generate alternative fall recovery strategies, complex motions with large contact areas such as crawling, or the generation of more autonomous locomotion and manipulation (loco-manipulation) actions (Fig.~\ref{fig:HitWithLeg}).  

Our goal is to go a step further in the understanding of the possible range of poses that can provide stability to the robot. To do so, we have been inspired by grasping. 
The main tools to understand how the hand can hold an object are the grasp taxonomies \cite{cutkosky_taxonomy_1989, feix_taxonomy_2009, Bernardin_sensor_2005, Kamakura_taxonomy_1989}. Grasp taxonomies have been proven to be useful in many contexts: to provide a benchmark to test the abilities of a new robotic hand, to simplify grasp synthesis, to guide autonomous grasp choices, or to inspire hand design, among others.

When a humanoid uses its body to gain stability through contacts with its environment, the dynamic equations to achieve equilibrium are similar to those of grasping, where the body plays the role of the hand and the object is the environment. %Therefore, we believe that a taxonomy of such poses could be as useful as grasping taxonomies are.
Let us refer to these body poses as {\it whole-body grasps}. In this work, we propose a classification of whole-body grasps exploring similar criteria as these used for taxonomies in grasping. Most grasping taxonomies define two main categories: precision and power grasping. In addition, Cutkosky classifies grasps according to object shape and tasks \cite{cutkosky_taxonomy_1989}. Kamakura according to task and the hand areas of contact (\cite{Kamakura_taxonomy_1989, Bernardin_sensor_2005})  and Feix et al. according to number/type of contacts and the configuration of the thumb \cite{feix_taxonomy_2009}. Indeed, there are many similar concepts. We can directly use the idea of precision vs. power grasping for whole-body grasps: poses that use contact with the torso vs. poses where contacts are realized only using the end-effectors. But also there is an important difference, almost all whole-body grasps are non-prehensile, {\it i.e.}, grasps that use the gravity to hold the object. We will explore in detail how this affects the criteria that used to obtain a whole-body pose taxonomy.

%nakaoka2004efficient - key poses, movement primitives to transfer from one motion to another.
%hoffmann2010body - body schema - talks about taxonomies on body representations
%
%assa2005action-selection of key poses to characterizee a motion
%
%chao2012human - characterization of poses from static drawings, sketches (to find motions in large databases of clips of motion capture)

Whole-body motion is widely studied in many areas such as computer graphics, biomechanics, and robotics. An important aspect of motion analysis is how to characterize motions, using sketches \cite{assa2005action} or smart selection of key frames \cite{kapadia_efficient_2013,nakaoka2004efficient}. A proper characterization of a motion is crucial for efficient search in whole-body motions large databases, but also for movement segmentation and semantic interpretations of actions with applications to imitation learning and autonomous robotics  \cite{Waechter2013,lieberman2004improvements,muhlig2010human}.
In the field of computer graphic animations, there are works that can successfully plan a motion with multiple contacts and complex interaction with the environment \cite{mordatch2012discovery,al2013relationship}. However, the autonomous generation of complex interactions with the environment for robotics is still an open problem due to the computation complexity of the planning, the robot lack of awareness of the available options and the uncertainty at the time of execution.

While in classic locomotion actions such as walking and running the transitions between double and single support poses are very well understood \cite{yin2007simbicon,kwon2005motion}, such transitions can become much more complex when the possibility of leaning against a surface with the hands is considered. We are interested in identifying balance poses during motions to be able to understand the motion and segment it into motion primitives based on the support poses. Our idea is then to reproduce and generate new locomotions that combine motion primitives based on support poses, similarly as it is done for manipulation actions \cite{muhlig2010human}.

% Also, several different classifications of body schema representations exist \cite{hoffmann2010body}, focusing on how the brain represents our body parts and their interaction with the environment. Although all the poses in our taxonomy modify the body schema, we focus on the classification of the possible poses more than in how to represent them.  In addition, 
However, the human body is complex, and so, from the biomechanics point of view, a taxonomy of stable human body poses could become very complex, as some humans have outstanding capabilities not only for stable walking, but for standing on a tip-toe and perform all kind of acrobatics movements. In this work, we focus on robotics and our main priority will be to simplify the large number of possible poses in a way that can be useful for current humanoid robots, such as HRP-2, HRP-4~\cite{kaneko2008humanoid}, ASIMO \cite{ASIMO1998}, TORO~\cite{TORO2014}, ARMAR-4~\cite{Asfour2013}, etc. Therefore, we do not intend to provide a complete taxonomy that covers all the possible configurations the human body affords.

To the best of our knowledge, there is no classification of static humanoid support poses. This is probably due to the relative novelty of the problem. 
A full taxonomy of whole-body grasps can have many interesting applications and uses, such as a tool for autonomous decision making, a guide to design complex motions combining different whole-body grasps, a way to simplify the control complexity,  a benchmark to test abilities for humanoid robots, and a way to improve recognition of body poses and transitions between them.

This paper is organized as follows. Section \ref{sec:Taxonomy} introduces the proposed taxonomy and discusses the criteria that have been taken into account. Section \ref{sec:Formalization} defines a support pose and its relationship with motion and actions, proposing a motion analysis method to identify the locomotion and manipualtion parts of an action. An example of the proposed analysis is shown in Section \ref{sec:ExampleMotionAnalysis}. Finally, Section \ref{sec:Conclusions} summarizes the presented work.

\section{TAXONOMY OF STABLE WHOLE-BODY POSES}\label{sec:Taxonomy}
When considering the whole body interacting with the environment, there is a wide range of different postures that the robot can adopt. We are interested in those poses that use  contacts for  balancing. Then, the limb end-effectors that are not used for balancing can be used to perform other manipulation tasks. This way, we provide a framework for loco-manipulation poses.

\begin{figure}[t]
\centering
\includegraphics[width=0.8\columnwidth]{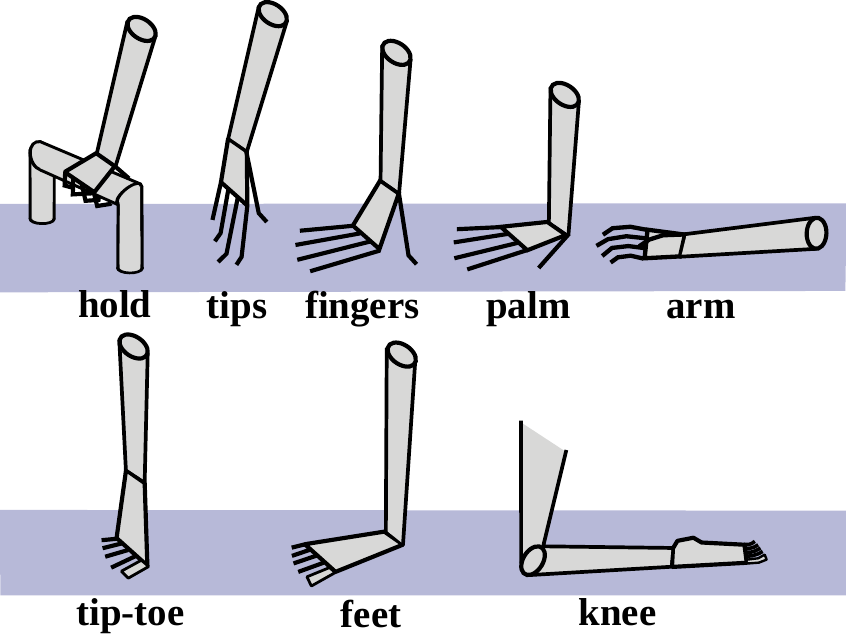}%
\caption{Types of support contacts with arms and legs.}
\label{fig:typesOfContact}%
\end{figure}

In Table \ref{fig:taxonomy} you can find our proposed taxonomy. It contains a total of 46 classes, divided into three main categories: standing, kneeling and resting. Each row corresponds to different number of supports, and in each row, different columns correspond to different contact types (see contact type legend at the bottom left corner or Table \ref{fig:taxonomy}). In addition, colors differentiate type of leg supports and poses under the gray area use line contacts (with arms or legs). 
The lines between boxes indicate possible pose transitions assuming only one change of support at a time.

%%%%%%%%%%%%%%%%%%%%%%%%%%%%%%%  TAXONOMY FIG %%%%%%%%%%%%%%%%%%%%%%%%%%%%%%%%
%\begin{sidewaysfigure*}
%\includegraphics[width=\textwidth]{figs/Taxonomy}%
%\caption{Proposed taxonomy for whole-body stability poses for loco-manipulation}
%\label{fig:taxonomy}%
%\end{sidewaysfigure*}

\afterpage{
 \begin{landscape}%\thispagestyle{empty}
\begin{table}%
\caption{Taxonomy of balancing whole-body poses}
\includegraphics[width=24cm]{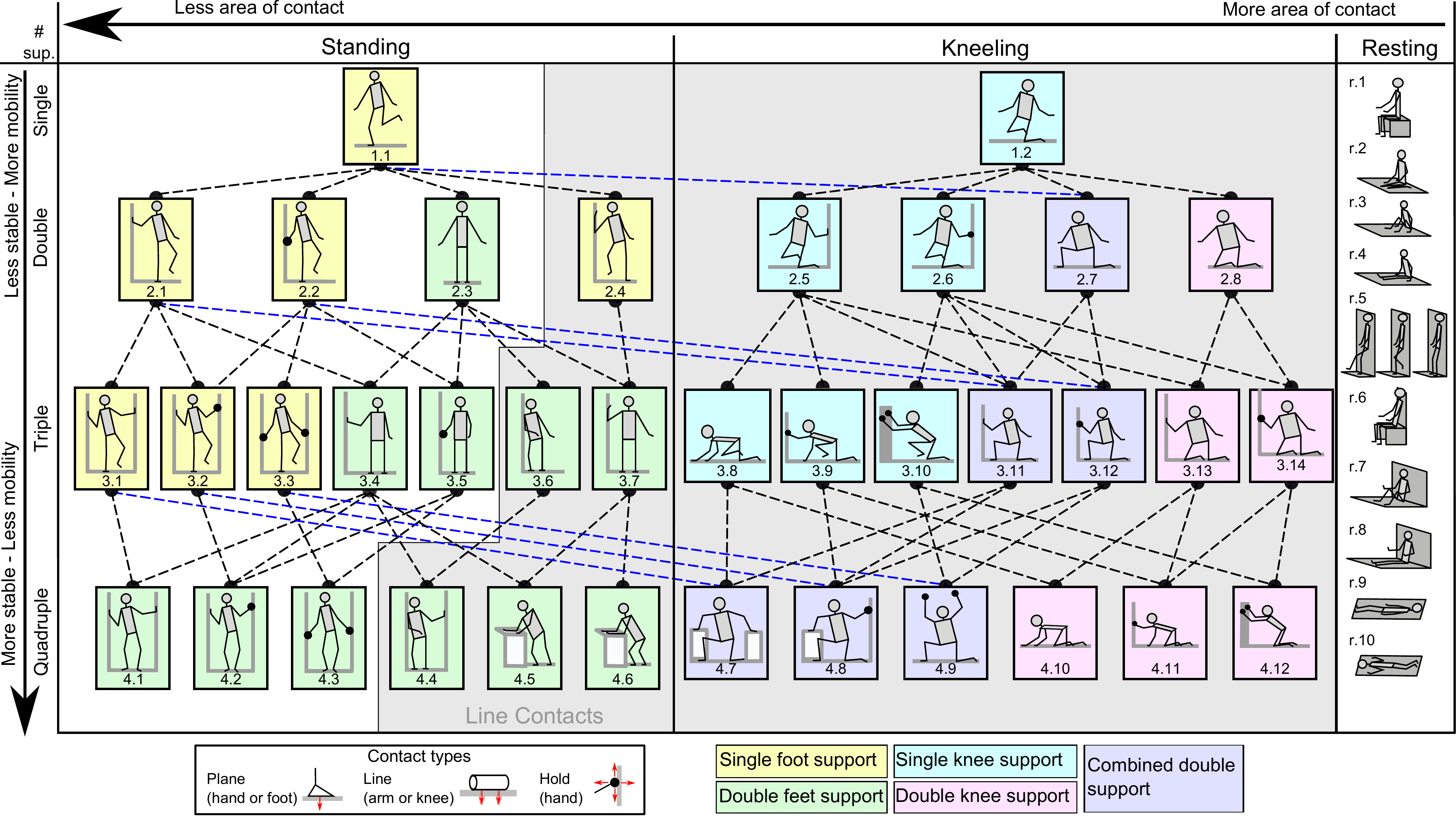}
The proposed taxonomy has 46 classes, including 18 standing poses, 18 kneeling poses and 10 resting configurations.  Sketches represent all the ranges of poses with the same number of supports and type of contact. Each class includes the symmetric cases when applicable. The lines provide possible road maps to transfer from one pose to another, assuming we can just perform one contact change at a time. Lines also provide a hierarchy among the poses. Blue lines represent transitions to different categories (from standing to kneeling).
\label{fig:taxonomy}
\end{table}
\end{landscape}
}

%%%%%%%%%%%%%%%%%%%%%%%%%%%%%%%%%%%%%%%%%%%%%%%%%%%%%%%%%%%%%%%%%%%%%%%%%%%%%%%%%%%%%%%%%%%%%%%%%%%%%%%%%%%%%%%%%%%%%%%%%%%%%%%%%%%%%%%%%%%%%%%%%%%%%%%%%%%%%%%%%%%%%
%%%%%%%%%%%%%%%%%%%%%%%%%%%%%%%%%%%%%%%%%%%%%%%%%%%%%%%%%%%%%%%%%%%%%%%%%%%%%%%%%%%%%%%%%%%%%%%%%%%%%%%%%%%%%%%%%%%%%%%%%%%%%%%%%%%%%%%%%%%%%%%%%%%%%%%%%%%%%%%%%%%%%
%%%%%%%%%%%%%%%%%%%%%%%%%%%%%%%%%%%%%%%%%%%%%%%%%%%%%%%%%%%%%%%%%%%%%%%%%%%%%%%%%%%%%%%%%%%%%%%%%%%%%%%%%%%%%%%%%%%%%%%%%%%%%%%%%%%%%%%%%%%%%%%%%%%%%%%%%%%%%%%%%%%%%
\subsection{Number and type of contacts}

One of the first relevant characteristics that greatly modifies the complexity of a motion is the number of supports with the environment. Kinematically, each support creates a new closed kinematic loop, and therefore, reduces by 1 the dimension of the feasible configuration space \cite{jaillet2013path,berenson2011task}. Dynamically, planning of complex motions tested on humanoid robots report higher execution times per higher number of supports \cite{saab_dynamic_2013}. Therefore, the number of supports with the environment is a clear first layer of classification, resulting in different rows for the taxonomy in Table \ref{fig:taxonomy}.

Contacts with environmental elements that do not provide support are not considered for the taxonomy classification.
%We consider only supports to use the taxonomy as a first level of classification of the support pose used to perform manipulation actions that may use other contacts with the environment. 
For instance, in Fig.\ref{fig:HitWithLeg} green marked contacts define the support pose, while the rest are contacts intended to manipulate the environment that do not affect at the support pose definition.

From the control point of view, it is very relevant the nature of the contact used to provide the support \cite{lemerle2008robust, mason2001mechanics} 
%Salisbury’s table of contact types, Mason book).  
and the part of the body that performs the support, because the resultant kinematics of the robot change accordingly. A fingertip contact is usually modeled as a point contact with friction, the foot support as plane contacts and arm leaning can be modeled using line with friction model \cite{mason2001mechanics}.
Fig. \ref{fig:typesOfContact} shows different possible types of support with the legs and arms. Considering all the types in the figure leads to 190 possible combinations.  Because we would like to keep our taxonomy simple, we have consider only 5 types: hold, palm, arm, feet, and knee support. These lead to the consideration of 51 combinations from which we have selected 36 (corresponding to the standing and kneeling poses). This choice has been done assuming that some combinations, while feasible, are not common. However, after further analysis of different motions is done, more classes may be included or excluded in the future.

% Finally, we only consider a kneeling contact where the sole of the feet are not in contact with the floor. If tip-toe contact is allowed, a more complex kneeling can be used where the foot is bended to allow floor contact at the tip toe and at the knee at the same time. This allows a direct transition from standing to kneeling  using a  movement where the center of mass is shifted until the robot falls into the kneeling pose or the standing one.  This has been implemented in the HRP-2 robot fall recovery \cite{kanehiro2003first}. 
% For this work, we only consider transitions with 1 contact change, and therefore, we have  left this option out of the taxonomy.
% 

%%%%%%%%%%%%%%%%%%%%%%%%%%%%%%%%%%%%%%%%%%%%%%%%%%%%%%%%%%%%%%%%%%%%%%%%%%%%%%%%%%%%%%%%%%%%%%%%%%%%%%%%%%%%%%%%%%%%%%%%%%%%%%%%%%%%%%%%%%%%%%%%%%%%%%%%%%%%%%%%%%%%%
%%%%%%%%%%%%%%%%%%%%%%%%%%%%%%%%%%%%%%%%%%%%%%%%%%%%%%%%%%%%%%%%%%%%%%%%%%%%%%%%%%%%%%%%%%%%%%%%%%%%%%%%%%%%%%%%%%%%%%%%%%%%%%%%%%%%%%%%%%%%%%%%%%%%%%%%%%%%%%%%%%%%%
%%%%%%%%%%%%%%%%%%%%%%%%%%%%%%%%%%%%%%%%%%%%%%%%%%%%%%%%%%%%%%%%%%%%%%%%%%%%%%%%%%%%%%%%%%%%%%%%%%%%%%%%%%%%%%%%%%%%%%%%%%%%%%%%%%%%%%%%%%%%%%%%%%%%%%%%%%%%%%%%%%%%%
\subsection{From precision to power whole-body grasps}

In addition to the standing and kneeling poses we have added 10 extra classes where there is contact with the torso.  We have called them resting poses. Poses from r.1 to r.4 are poses where still balance needs to be achieved, but the inclination of the torso needs to be controlled. Poses from r.5 to r.6 are stable provided that the areas of contact are flat and with friction. Finally, using poses from r.7 to r.10 the robot is unlikely to lose balance and can be considered safe and completely in rest, but with very limited mobility.

%The torso usually contains most of the mass of the robot, and therefore, it can modify greatly the conditions for equilibrium. The corresponding grasping poses are those using the palm, that is, power grasps, and are usually used to achieve  stable grasps. Similarly, the more area of the torso in contact, the more stable the pose is, begin completely stable in laying down poses such as r.9 and r.10. 

% The poses at the left side of the Table \ref{fig:taxonomy} use the extremities end-effectors to balance. Thus, when compared to grasping, they correspond to precision grasps. All poses with line contacts (gray area) correspond to intermediate grasps. Finally, contact with the torso correspond to power grasps \cite{feix_taxonomy_2009}. In a  similar way as precision and power grasping constitute a trade-off between dexterity and stability, end-effector in contact and torso-in-contact poses constitute a trade-off between mobility and stability.  

At this stage of work, no transitions are shown between resting poses and the rest of the table. Such transitions are more complex and require further motion analysis that will be left for future work.

\subsection{Shape of the environment}

Many grasping taxonomies include the shape of the object as a criteria for grasp choice. Indeed, object shape and size have a great influence on the ability for grasping and manipulation.

\begin{figure}[b]
\includegraphics[width=\columnwidth]{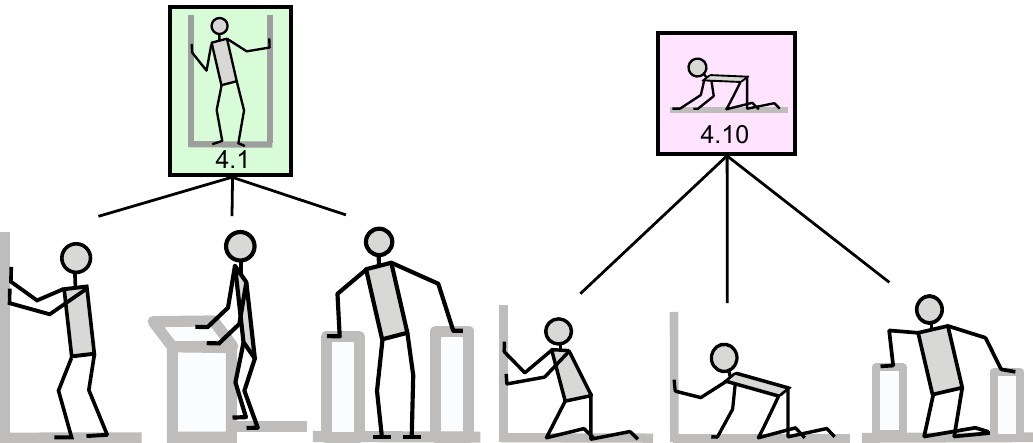}%
\caption{Implementation of poses of the classes 4.1 and 4.10 for different environment shapes.}%
\label{fig:onShapeEnviornment}%
\end{figure}

However,  there is a fundamental difference between hand grasping and whole-body grasping: the need of gravity to reach force closure. A hand grasp will always start with no contacts at all, and after grasping, it may or may not start a manipulation motion that can be maintaining constant contacts (in-Grasp Manipulation) or performing re-grasps \cite{ma2011dexterity}. On the contrary, a whole-body grasp is always part of a motion sequence of re-grasps that will always start with at least one contact with the environment (even if one of the phases has no contact as in a running locomotion or jumping).

For this reason, we believe that whole-body grasp choice will not depend as much on the shape of the room, but on the task/motion the pose occurs. For instance, given the two poses in Fig. \ref{fig:onShapeEnviornment}, if the environment is only a floor and a wall, as in the left lower shape, the choice between pose $4.1$ or $4.10$ will be mainly influenced by the motion/task. For instance, at the beginning of a fall recovery motion the choice will most likely be $4.10$ intercalated with $3.13$, while at the end, once the robot is already standing, the choice will most probably be $4.1$. 
  
In addition, the number and nature of constraints the poses in a class have to maintain are all the same regardless of environment shape, just the directions of the normals on the contact points change. 
%Therefore, the equilibrium equations are the same, topologically and computationally, and so are the controllers.

% While it is not exactly the same to lean on a table or against a wall, we have decided to consider them as the same class because the equilibrium equations are mainly the same, topologically and computationally,
% and most relevant differences are mainly influenced by the type of contact. 

For these reasons, we consider poses with different environment shape to be the same class. However, environment shape has to be taken into account. Therefore, in Section \ref{sec:Formalization} we introduce a definition for pose class that in each instantiation contains information about the specific shape of the environment using the normals at the contacts.  

% number of available surfaces to support. In the case of holding, the existence of the handle in the scene can allow the choice of a pose that includes handling, and the shape of the handle will only affect at the chosen grasp to perform the hold, which falls into the scope of grasping.

%%%%%%%%%%%%%%%%%%%%%%%%%%%%%%%%%%%%%%%%%%%%%%%%%%%%%%%%%%%%%%%%%%%%%%%%%%%%%%%%%%%%%%%%%%%%%%%%%%%%%%%%%%%%%%%%%%%%%%%%%%%%%%%%%%%%%%%%%%%%%%%%%%%%%%%%%%%%%%%%%%%%%
%%%%%%%%%%%%%%%%%%%%%%%%%%%%%%%%%%%%%%%%%%%%%%%%%%%%%%%%%%%%%%%%%%%%%%%%%%%%%%%%%%%%%%%%%%%%%%%%%%%%%%%%%%%%%%%%%%%%%%%%%%%%%%%%%%%%%%%%%%%%%%%%%%%%%%%%%%%%%%%%%%%%%
%%%%%%%%%%%%%%%%%%%%%%%%%%%%%%%%%%%%%%%%%%%%%%%%%%%%%%%%%%%%%%%%%%%%%%%%%%%%%%%%%%%%%%%%%%%%%%%%%%%%%%%%%%%%%%%%%%%%%%%%%%%%%%%%%%%%%%%%%%%%%%%%%%%%%%%%%%%%%%%%%%%%%
\subsection{Stability}

%It is not the scope of this work to show a classification of whole-body poses according to a measure of stability. 
The taxonomy in Table \ref{fig:taxonomy} is organized so that the less stable poses, with less number of supports lie on the upper left side, while the most stable ones on the lower right side, assuming that the more number of contacts and the larger the surfaces of contact, the more stable the robot is. Works like \cite{huang2011tradeoff}  show that there is a trade-off between stability and maneuverability during a goal-directed whole-body movements. In the taxonomy, we observe a similar tradeoff with mobility vs. stability.

However, inside any class it is possible to obtain different levels of stability depending on the support region \cite{bretl2008testing} and the sum of the contact wrenches \cite{wieber_stability_2002}.

\section{CLASSIFICATION OF ACTIONS AND MOTIONS}\label{sec:Formalization}
The proposed taxonomy induces a formalization of whole-body motions/actions depending on the support poses. To formalize this in more detail, we first need to define how to instantiate a pose class and the relationship between poses and motions. 

\subsection{Support pose class}
In this context, we first need to define a contact as
\begin{equation}
C = \{l, m, {\bf c}, {\bf n}\}
\end{equation}
where $l$ is the link in contact, $m$ is the model of the contact, ${\bf c}$ are the global coordinates of the contact location and ${\bf n}$ the normal direction of the surface of contact. The contact model defines the number and nature of constraints, which will be unilateral in the case of a plane or a line contact, or bilateral in the case of a hold support. Robots can autonomously obtain the information about the shape of the environment using advanced perception techniques. Our group is working on methods to extract geometric primitives of the environment, providing information about location and normal direction of possible contacts in the scene \cite{Kaiser2014Extracting}.

Then, the instantiation of a class of the taxonomy is defined by
\begin{equation}
 \{ id, {\bf p}, \mathcal{C}= \{C_i, i=1\dots m\}, \mathcal{N} \}
\end{equation}
where $id$ is an identification of the class in the taxonomy, $p$ the location of the center of mass of the robot (CoM), $\mathcal{C}$ the set of $m$ contacts and $\mathcal{N}$ the set of neighbor classes.

This way, for each class the robot can be represented as a simplified model that contains the CoM, contact locations and contact normal directions in a similar way as in the works of Lemerle group \cite{lemerle2011new, lemerle2008robust}. The information that defines each class instantiation is sufficient to define the set of constraints and equilibrium conditions that need to be satisfied \cite{bretl2008testing}, and to design the corresponding controllers to allow motions inside the class.

\subsection{Pose transitions and motions}
A transition between two classes can happen by first imposing the constraints of the current and destination class, and then shifting to only the constraints of the destination class. This induces the definition of two types of motions
\begin{enumerate}
 \item \textbf{Inside class motion}: A purely manipulation action will happen inside a single class. It includes other manipulation motions and therefore, extra contacts with objects, always with the objective of manipulation. As a manipulation motion, it can be semantically segmented and interpreted as done in \cite{Waechter2013}.
 \item \textbf{Transition class motion}: motions that define a transition between poses. The motion still occurs inside a class, but the motion consists in the shifting towards a destination class, as part of a locomotion. For instance, a double feet support motion that shifts towards a right foot support ($2.3\rightarrow 1.1$).
\end{enumerate}

Note that both motions happen always inside the same support class, but in the second case, the destination class is relevant for the motion definition.

In this context, the taxonomy provides a classification that allows to store previously computed motions associated to support poses. Transition class motions can be simple enough to convert them to motion primitives. Inside class motions may need further segmentation. Next, we go a step further to associate motion to tasks or actions.

%%%%%%%%%%%%%%%%%%%%%%%%%%%%%%%%%%%%%%%%%%%%%%%%%%%%%%%%%%%%%%%%%%%%%%%%%%%%%%%%%%%%%%%%%%%%%%%%%%%%%%%%%%%%%%%%%%%%%%%%%%%%%%%%%%%%%%%%%%%%%%%%%%%%%%%%%%%%%%%%%%%%%
%%%%%%%%%%%%%%%%%%%%%%%%%%%%%%%%%%%%%%%%%%%%%%%%%%%%%%%%%%%%%%%%%%%%%%%%%%%%%%%%%%%%%%%%%%%%%%%%%%%%%%%%%%%%%%%%%%%%%%%%%%%%%%%%%%%%%%%%%%%%%%%%%%%%%%%%%%%%%%%%%%%%%
%%%%%%%%%%%%%%%%%%%%%%%%%%%%%%%%%%%%%%%%%%%%%%%%%%%%%%%%%%%%%%%%%%%%%%%%%%%%%%%%%%%%%%%%%%%%%%%%%%%%%%%%%%%%%%%%%%%%%%%%%%%%%%%%%%%%%%%%%%%%%%%%%%%%%%%%%%%%%%%%%%%%%
\subsection{Tasks and actions}\label{subsec:actions}
 Previous works on manipulation tasks such as \cite{worgotter_simple_2013} defined an action as an interaction between a hand and an object to induce a change at the object. In addition, they defined action components as those time-points where contact relations between hand and objects change. With these simple rules, they can classify all single hand manipulations in only 6 categories. 

Following a similar approach,  we can also define a whole-body action as an interaction between the body and the environment, but with two possible objectives: to induce a change in the environment or in the body itself. In the latter, we refer to whole-body actions where the main objective is to relocate (locomotion) or to gain stability (balance), not to change the environment. Action components can also be defined as those time-points where the number of contacts change. 

Using these definitions, we can define three main categories of actions associated to the motions we defined before: 
\begin{itemize}
	\item (I) actions that are intended to change the environment (that will be done using inside class motions) 
	\item (II) actions that are intended to change the body/robot (using transition class motions) and
	\item (III) a combination of the above: actions where supports are used both to balance and to change the environment.
\end{itemize}

\begin{figure}[bt]
\includegraphics[width=\columnwidth]{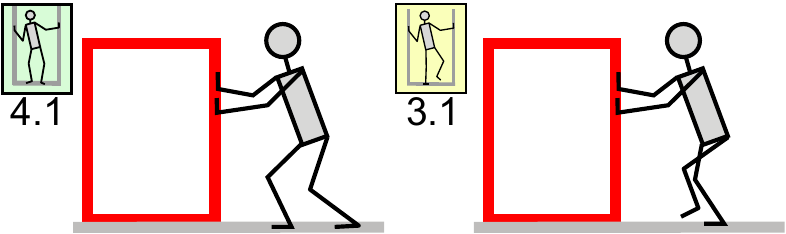}%
\caption{Implementation of poses of the class 4.1 and 3.1, for a type III manipulation where all contacts are used both for balance and to change the environment.}%
\label{fig:pushBigObject}%
\end{figure}

Fig. \ref{fig:HitWithLeg} shows an example of the action of type (I) called in \cite{worgotter_simple_2013} {\em hit object}. The task requires a single end-effector to hit an object. Then, according to the requirements of the object (location, size or weight), we could decide between 1, 2 or 3 support contacts to define the support pose where the manipulation can take place. In Fig. \ref{fig:HitWithLeg} we show 4 possible support poses: the standing pose 2.3, or pose 3.4 using an extra support on a table to be able to reach further, or a pose where the manipulation is done with the foot and the support is provided using the hands as in poses 3.1 or 2.1. 

Special cases are actions of type (III), like the example in Fig. \ref{fig:pushBigObject}. Here, the four contacts are used both for support and to modify the environment. We can identify such actions because they start as manipulation actions (inside class motions) but they occur during several pose transitions, and therefore, constitute a combined locomotion and manipulation action.

This framework induces a set of segmentation criteria for a given motion that, provided that we can differentiate support contacts and manipulation contacts, subdivides a motion into pieces that can be related with types of actions. For actions identified as manipulation (type I), further segmentation based on the manipulation contacts can be performed \cite{Waechter2013}, providing a hierarchy of segments distinguishing between the locomotion and the manipulation parts of an action.

The lines between boxes in the taxonomy represent possible transitions of only one support change. 
However, we want to analyze support pose transitions during loco-manipulation motions in humans to improve the proposed taxonomy and validate the proposed transitions. In addition, the motion analysis can provide a better semantic understanding of complex locomotion and manipulation actions for imitation learning and autonomous decision making applications. 

To do so, we propose to use the publicly available KIT whole-body human motion database \cite{KITDatabase, Mandery2015}. The database contains many motion capture data not only of human motions, but also from objects that are being used or manipulated during the motion. Including environmental elements in the motion capture data is the key aspect that allows us to use this database to analyze loco-manipulation actions. In addition, the database provides models of the objects and a normalized subject-independent representation of the motion capture data, based on a framework called Master Motor Map (MMM) \cite{Terlemez2014_MMM}. 
The idea behind the MMM framework is to offer a unifying reference model of the human body with kinematic and dynamic parameters. This allows us to analyze the position data of different segments of the body, and to detect collision with the objects of the environment. In addition, MMM motions can be converted to other robots with different kinematic and dynamic structures, offer the possibility to transfer the pose transition motions to a humanoid robotic platform.

The atoms of motion, segmented using the proposed analysis, can be use to define Dynamic Motion Primitives \cite{ijspeert2002movement}. The taxonomy can help to define a formal grammar to build new motions to provide robots with autonomous techniques to generate possible motions for a given environment.

In the next section we show an example of the proposed motion analysis to one of the motions in the database. The analysis allows us to visualize the motion as a subgraph of the taxonomy.

\begin{figure}[tb]
\includegraphics[width= \columnwidth ]{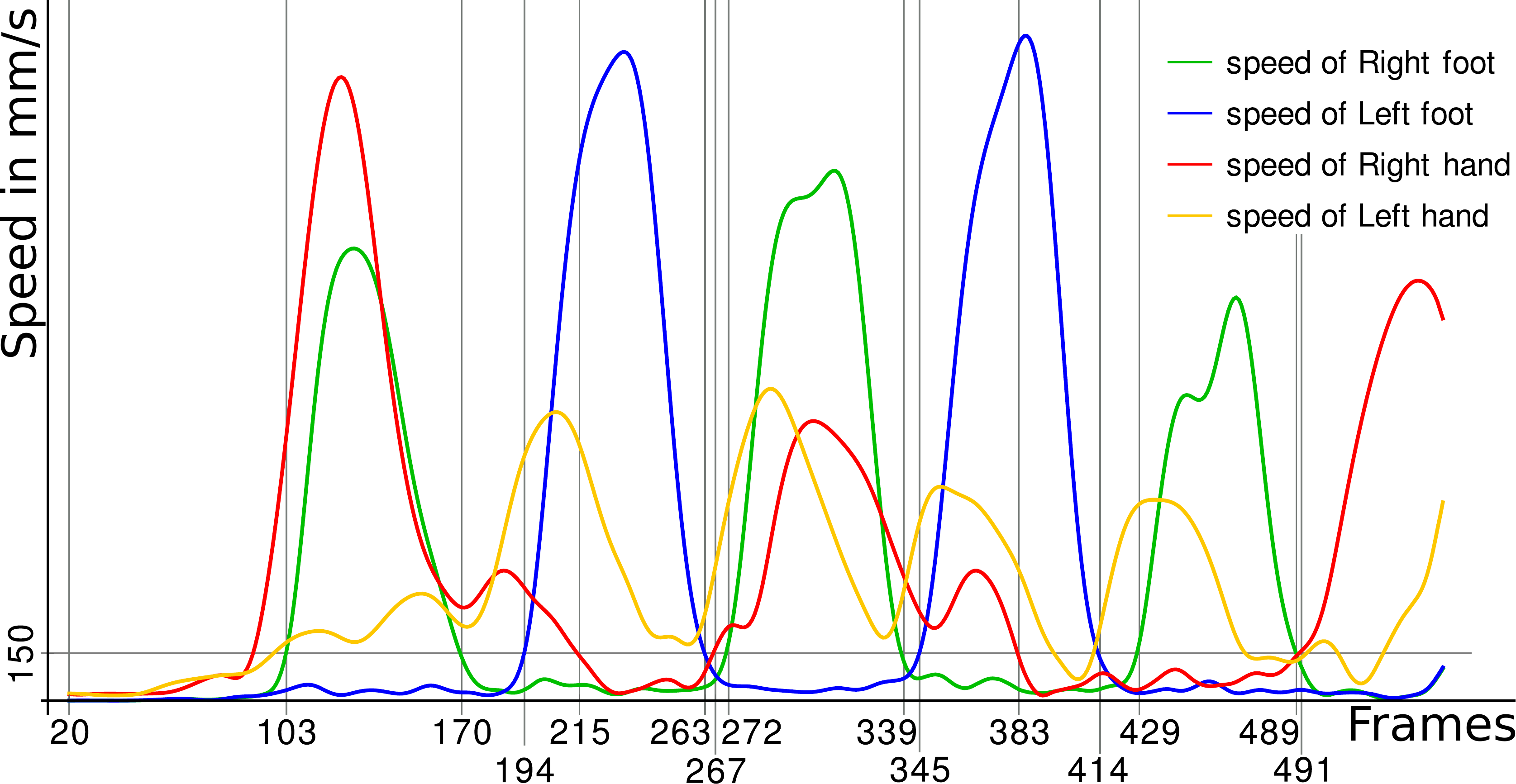}%
\caption{Speed of the considered end-effectors during the motion. Grid lines on the X axes show the segmentation frames detected where there was a change on support contacts. Grid lines on the Y axes show the speed threshold considered. }%
\label{fig:speeds}%
\end{figure}

\section{EXAMPLE}\label{sec:ExampleMotionAnalysis}

\begin{figure*}[bt]
\centering
\includegraphics[width= \linewidth ]{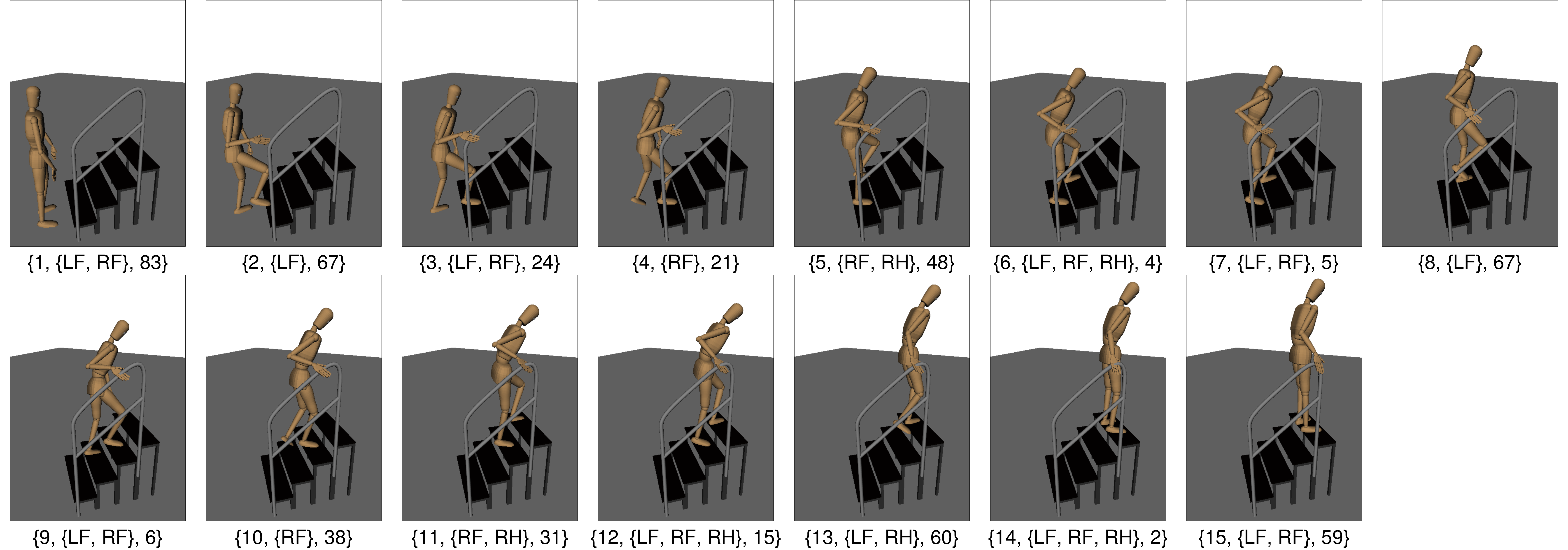}%
\caption{Result of the motion segmentation. We show the middle frame of each segment, with a label that indicates the order, the extremities in contact and the length of the segment in number of frames. RF and LF correspond to right and left foot, while RH and LH right and left hand.}%
\label{fig:framesOfSegmentation}%
\end{figure*}

The analyzed motion consists of going up a set of stairs with a handle at the right hand side, lasting 5.5s with 100 frames/s \footnote{Raw data files can be found in upstairs05 motion files in https://motion-database.humanoids.kit.edu/details/motions/383/}. We combine a segmentation based on collision detection to detect contacts, that has been used in previous works to semantically interpret manipulation actions \cite{Waechter2013} with an analysis of the velocities of the extremities that can be in contact. A support contact is characterized by having zero 6-dim velocity \cite{sentis_2010_compliant_Springer}, however, the motion capture systems are noisy and they cannot capture the exact contact point location, and therefore, the velocities of the support segment are only close to zero. We have simplified to consider only the module of the positional velocity. The velocities are obtained by numerical differentiation of the position data corresponding to the four end-effectors considered for possible supports: the two feet and the two hands. After inspecting the frequency content of these signals, we have low-pass filtered the position data at a cut-off frequency of 1.5Hz to reduce the effect of noise. The resulting computed velocities are plotted in Fig. \ref{fig:speeds} for each of the end-effectors.

Despite the noise in the data, we can identify different hypotheses of support using a speed threshold of $0.15 m/s$. Such hypotheses are then validated against the contact segmentation that is computed based on collision detection between the objects of the scene (the stairs and the floor) \cite{Waechter2013}. We discard the hypothesis of support if the extremity is not in contact with any object of the environment. The velocity information is considered more relevant in this case than the collision, because often the hand is very close to the handle but slides on it, without providing a real support. 

In Fig. \ref{fig:framesOfSegmentation} the resulting segments are represented showing the middle frame of each segment, and they are labeled according to the detected supports (RF and LF stand for left and right foot, while RH and LH stand for right/left hand). We also show the length of each segment as number of frames.

\begin{figure}[bt]
\centering
\includegraphics[width= 0.9\columnwidth ]{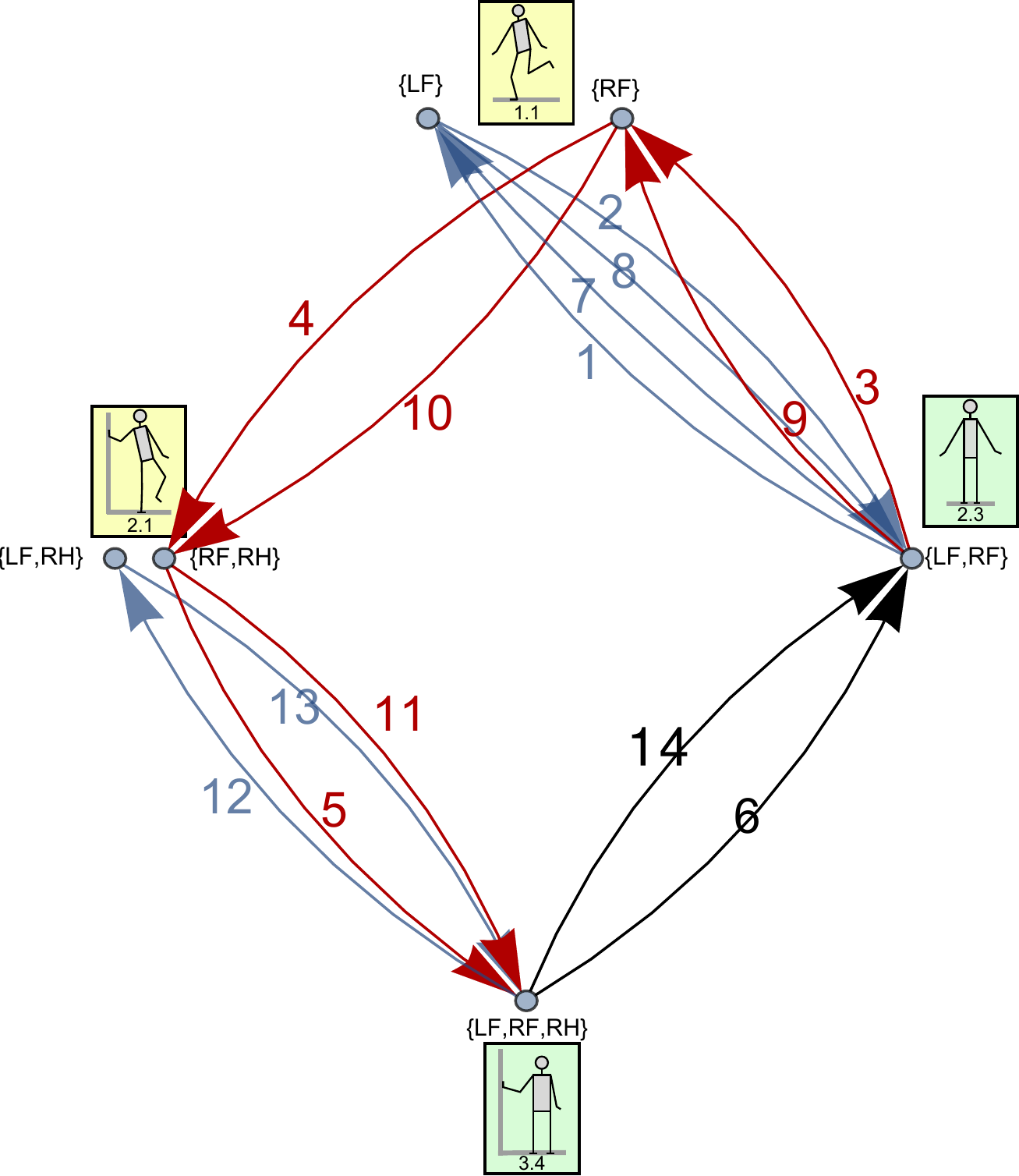}%
\caption{Graph analysis of the pose transition occurring in the analyzed motion. Blue edges correspond to transitions to or from left foots, and red edges transitions to or from right foots. The edge labels indicate the order of the transition, in accordance to the labeling in Fig. \ref{fig:framesOfSegmentation}}.%
\label{fig:motionGraph}%
\end{figure}

Using the taxonomy of whole-body poses, we can analyze further the obtained segmentation by showing the number of pose transitions occurring during the motion as a graph (Fig. \ref{fig:motionGraph}). Each arrow shows the direction between origin and destination pose, and are labeled according to the order they occur. This offers a visual representation that allows us to rapidly observe that the motion starts and finishes in the double feet support and that the single support pose is visited 4 times, 2 for each foot (and so, the motion contains 4 steps). We can also observe that the cycles are not symmetric, because the handle is on the right side of the body. Therefore the support on the handle occur mostly during the right foot support, while during the left foot support both the right foot and the right hand are swung towards the next support location. The left foot with hand support occurs at the end of the motion, when both feet go to the same step to finalize the walking.

It is also worth noticing that some of the segments are very short (5 to 2 frames). This implies that in practice, more than one change of support occurs almost immediately. However, we believe that for robotic movements, is safer to always take into account this intermediate short poses.
In walking motions, the double support phase is always much shorter than the single support, and the faster the motion, the shorter the double support. In this case, we can also say that beside the initial and final segment, the double foot support with or without hand have much shorter segment duration than phases with single foot support (with or without hand support).

As a future work, we plan to record new motions involving more poses of the taxonomy, and combining actions of locomotion and manipulation. Analyzing the motion data will allow us to validate the suggested taxonomy, propose different routings of transitions, and modify the taxonomy accordingly if necessary.

%\section{APPLICATIONS}\label{sec:Applications}
%\input{sections/Applications}

\section{CONCLUSIONS}\label{sec:Conclusions}
This work, for the first time, has proposed a taxonomy of whole-body balancing poses containing 46 classes, divided into three main categories, considering number and type of support and possible transitions between support poses. We have analyzed known grasping criteria used to classify robot grasps, but focusing on the demands of whole-body poses. As opposed to grasping, we have given less relevance to environment shape and more to the type of contact the body uses to provide a support pose. 

We have proposed a formal definition to characterize a pose, as well as a characterization of motions according to the class pose where they take place. Analyzing an example of whole-body motion of going upstairs, we have shown a segmentation technique that allows us to identify several poses of the taxonomy, segment the motion according to the identified poses and provide a graph representation of a motion as transitions between such poses.

We believe the proposed taxonomy has a lot of potential to be used in many areas of humanoid robotics. In  future work, we plan to refine our classification using further motion data, store classified motions according to pose and convert and adapt such motions to humanoid robotic platforms that use their environment for extra support. Nonetheless, the range of applications is certainly much wider and we expect to contribute in many aspects of humanoid robotics research in the upcomming years.

\section*{ACKNOWLEDGMENT}

The research leading to these results has received funding from the European Union Seventh Framework Programme under grant agreement no. 611832 (WALK-MAN).

The authors would like to thank \"Omer Terlemez, Mirko W\"achter and Christian Mandery for their collaboration and fruitful discussions about this work.

%%%%%%%%%%%%%%%%%%%%%%%%%%%%%%%%%%%%%%%%%%%%%%%%%%%%%%%%%%%%%%%%%%%%%%%%%%%%%%%%

%\bibliographystyle{IEEEtran}

%\bibliography{./bibtexFiles/biblioAffordances,./bibtexFiles/biblioWalking,./bibtexFiles/biblioGrasping,./bibtexFiles/biblioWholeBody}

\begin{thebibliography}{99}

\bibitem{kajita_biped_2010}
S.~Kajita, M.~Morisawa, K.~Miura, S.~Nakaoka, K.~Harada, K.~Kaneko,
  F.~Kanehiro, and K.~Yokoi, ``Biped walking stabilization based on linear
  inverted pendulum tracking,'' in \emph{{IEEE/RSJ} International Conference on
  Intelligent Robots and Systems}, 2010, p. 4489–4496.

\bibitem{pratt_capture_2006}
J.~Pratt, J.~Carff, S.~Drakunov, and A.~Goswami, ``Capture point: A step toward
  humanoid push recovery,'' in \emph{6th {IEEE-RAS} International Conference on
  Humanoid Robots}, 2006, p. 200–207.

\bibitem{wieber_stability_2002}
P.-B. Wieber, ``On the stability of walking systems,'' in \emph{Proceedings of
  the international workshop on humanoid and human friendly robotics}, 2002.

\bibitem{sentis_2010_compliant_TRO}
L.~Sentis, J.~Park, and O.~Khatib, ``Compliant control of multicontact and
  center-of-mass behaviors in humanoid robots,'' \emph{{IEEE} Transactions on
  Robotics}, vol.~26, no.~3, pp. 483--501, 2010.

\bibitem{saab_dynamic_2013}
L.~Saab, O.~E. Ramos, F.~Keith, N.~Mansard, P.~Soueres, and J.-Y. Fourquet,
  ``Dynamic whole-body motion generation under rigid contacts and other
  unilateral constraints,'' \emph{{IEEE} Transactions on Robotics}, vol.~29,
  no.~2, pp. 346--362, 2013-04.

\bibitem{lengagne_2013_generation}
S.~Lengagne, J.~Vaillant, E.~Yoshida, and A.~Kheddar, ``Generation of
  whole-body optimal dynamic multi-contact motions,'' \emph{The International
  Journal of Robotics Research}, vol.~32, no.~9, pp. 1104--1119, 2013.

\bibitem{Hirukawa_2006_AdiosZMP}
H.~Hirukawa, S.~Hattori, K.~Harada, S.~Kajita, K.~Kaneko, F.~Kanehiro,
  K.~Fujiwara, and M.~Morisawa, ``A universal stability criterion of the foot
  contact of legged robots- adios {ZMP},'' in \emph{Proceedings of the IEEE
  International Conference on Robotics and Automation}, 2006, pp. 1976--1983.

\bibitem{khatib_suprapeds_2014}
O.~Khatib and S.-Y. Chung, ``{SupraPeds}: Humanoid contact-supported locomotion
  for {3D} unstructured environments,'' in \emph{Proceedings of the IEEE
  International Conference on Robotics and Automation}, 2014, pp. 2319--2325.

\bibitem{cutkosky_taxonomy_1989}
M.~R. Cutkosky, ``On grasp choice, grasp models, and the design of hands for
  manufacturing tasks,'' \emph{Robotics and Automation, {IEEE} Transactions
  on}, vol.~5, no.~3, p. 269–279, 1989.

\bibitem{feix_taxonomy_2009}
T.~Feix, R.~Pawlik, H.-B. Schmiedmayer, J.~Romero, and D.~Kragic, ``A
  comprehensive grasp taxonomy,'' in \emph{Robotics, Science and Systems:
  Workshop on Understanding the Human Hand for Advancing Robotic Manipulation},
  2009.

\bibitem{Bernardin_sensor_2005}
K.~Bernardin, K.~Ogawara, K.~Ikeuchi, and R.~Dillmann, ``A sensor fusion
  approach for recognizing continuous human grasping sequences using hidden
  markov models,'' \emph{{IEEE} Transactions on Robotics}, vol.~21, no.~1, pp.
  47--57, 2005.

\bibitem{Kamakura_taxonomy_1989}
N.~Kamakura, \emph{Te no katachi Te no ugoki}.\hskip 1em plus 0.5em minus
  0.4em\relax Tokyo, Japan: Ishiyaku, 1989.

\bibitem{assa2005action}
J.~Assa, Y.~Caspi, and D.~Cohen-Or, ``Action synopsis: pose selection and
  illustration,'' \emph{ACM Transactions on Graphics (TOG)}, vol.~24, no.~3,
  pp. 667--676, 2005.

\bibitem{kapadia_efficient_2013}
M.~Kapadia, I.-k. Chiang, T.~Thomas, N.~I. Badler, J.~T. Kider~Jr, and
  {others}, ``Efficient motion retrieval in large motion databases,'' in
  \emph{Proceedings of the {ACM} {SIGGRAPH} Symposium on Interactive 3D
  Graphics and Games}.\hskip 1em plus 0.5em minus 0.4em\relax {ACM}, 2013, p.
  19–28.

\bibitem{nakaoka2004efficient}
S.~Nakaoka, A.~Nakazawa, and K.~Ikeuchi, ``An efficient method for composing
  whole body motions of a humanoid robot,'' in \emph{Proceedings of the Tenth
  International Conference on Virtual Systems and Multimedia (VSMM)}, 2004, pp.
  1142--1151.

\bibitem{Waechter2013}
M.~W\"achter, S.~Schulz, T.~Asfour, E.~Aksoy, F.~W\"org\"otter, and
  R.~Dillmann, ``Action sequence reproduction based on automatic segmentation
  and object-action complexes,'' in \emph{IEEE/RAS International Conference on
  Humanoid Robots (Humanoids)}, 2013, pp. 189--195.

\bibitem{lieberman2004improvements}
J.~Lieberman and C.~Breazeal, ``Improvements on action parsing and action
  interpolation for learning through demonstration,'' in \emph{Humanoid Robots,
  2004 4th IEEE/RAS International Conference on}, vol.~1.\hskip 1em plus 0.5em
  minus 0.4em\relax IEEE, 2004, pp. 342--365.

\bibitem{muhlig2010human}
M.~Muhlig, M.~Gienger, and J.~J. Steil, ``Human-robot interaction for learning
  and adaptation of object movements,'' in \emph{Intelligent Robots and Systems
  (IROS), 2010 IEEE/RSJ International Conference on}.\hskip 1em plus 0.5em
  minus 0.4em\relax IEEE, 2010, pp. 4901--4907.

\bibitem{mordatch2012discovery}
I.~Mordatch, E.~Todorov, and Z.~Popovi{\'c}, ``Discovery of complex behaviors
  through contact-invariant optimization,'' \emph{ACM Transactions on Graphics
  (TOG)}, vol.~31, no.~4, p.~43, 2012.

\bibitem{al2013relationship}
R.~A. Al-Asqhar, T.~Komura, and M.~G. Choi, ``Relationship descriptors for
  interactive motion adaptation,'' in \emph{Proceedings of the 12th ACM
  SIGGRAPH/Eurographics Symposium on Computer Animation}.\hskip 1em plus 0.5em
  minus 0.4em\relax ACM, 2013, pp. 45--53.

\bibitem{yin2007simbicon}
K.~Yin, K.~Loken, and M.~van~de Panne, ``Simbicon: Simple biped locomotion
  control,'' \emph{ACM Transactions on Graphics}, vol.~26, no.~3, pp. 105--1 --
  105--10, 2007.

\bibitem{kwon2005motion}
T.~Kwon and S.~Y. Shin, ``Motion modeling for on-line locomotion synthesis,''
  in \emph{Proceedings of the 2005 ACM SIGGRAPH/Eurographics symposium on
  Computer animation}.\hskip 1em plus 0.5em minus 0.4em\relax ACM, 2005, pp.
  29--38.

\bibitem{kaneko2008humanoid}
K.~Kaneko, K.~Harada, F.~Kanehiro, G.~Miyamori, and K.~Akachi, ``Humanoid robot
  {HRP}-3,'' in \emph{Intelligent Robots and Systems, 2008. IROS 2008. IEEE/RSJ
  International Conference on}.\hskip 1em plus 0.5em minus 0.4em\relax IEEE,
  2008, pp. 2471--2478.

\bibitem{ASIMO1998}
K.~Hirai, M.~Hirose, Y.~Haikawa, and T.~Takenaka, ``The development of {H}onda
  humanoid robot,'' in \emph{Robotics and Automation, 1998. Proceedings. 1998
  IEEE International Conference on}, vol.~2.\hskip 1em plus 0.5em minus
  0.4em\relax IEEE, 1998, pp. 1321--1326.

\bibitem{TORO2014}
J.~Englsberger, A.~Werner, C.~Ott, B.~Henze, M.~A. Roa, G.~Garofalo, R.~Burger,
  A.~Beyer, O.~Eiberger, K.~Schmid, and A.~Albu-Schäffer, ``Overview of the
  torque-controlled humanoid robot {TORO},'' in \emph{IEEE-RAS International
  Conference on Humanoid Robots}, 2014.

\bibitem{Asfour2013}
T.~Asfour, J.~Schill, H.~Peters, C.~Klas, J.~B\"ucker, C.~Sander, S.~Schulz,
  A.~Kargov, T.~Werner, and V.~Bartenbach, ``{ARMAR-4: A 63 DOF Torque
  Controlled Humanoid Robot},'' in \emph{{IEEE-RAS} International Conference on
  Humanoid Robots}, Atlanta, USA, October 2013, pp.~--.

\bibitem{jaillet2013path}
L.~Jaillet and J.~M. Porta, ``Path planning under kinematic constraints by
  rapidly exploring manifolds,'' \emph{Robotics, IEEE Transactions on},
  vol.~29, no.~1, pp. 105--117, 2013.

\bibitem{berenson2011task}
D.~Berenson, S.~S. Srinivasa, and J.~Kuffner, ``Task space regions: A framework
  for pose-constrained manipulation planning,'' \emph{The International Journal
  of Robotics Research}, vol.~30, no.~12, pp. 1435--1460, 2011.

\bibitem{lemerle2008robust}
C.~Collette, A.~Micaelli, C.~Andriot, and P.~Lemerle, ``Robust balance
  optimization control of humanoid robots with multiple non coplanar grasps and
  frictional contacts,'' in \emph{Robotics and Automation, 2008. ICRA 2008.
  IEEE International Conference on}.\hskip 1em plus 0.5em minus 0.4em\relax
  IEEE, 2008, pp. 3187--3193.

\bibitem{mason2001mechanics}
M.~T. Mason, \emph{Mechanics of robotic manipulation}.\hskip 1em plus 0.5em
  minus 0.4em\relax MIT press, 2001.

\bibitem{ma2011dexterity}
R.~R. Ma and A.~M. Dollar, ``On dexterity and dexterous manipulation,'' in
  \emph{Advanced Robotics (ICAR), 2011 15th International Conference on}.\hskip
  1em plus 0.5em minus 0.4em\relax IEEE, 2011, pp. 1--7.

\bibitem{huang2011tradeoff}
H.~J. Huang and A.~A. Ahmed, ``Tradeoff between stability and maneuverability
  during whole-body movements,'' \emph{PLoS One}, vol.~6, no.~7, p. e21815,
  2011.

\bibitem{bretl2008testing}
T.~Bretl and S.~Lall, ``Testing static equilibrium for legged robots,''
  \emph{Robotics, IEEE Transactions on}, vol.~24, no.~4, pp. 794--807, 2008.

\bibitem{Kaiser2014Extracting}
P.~Kaiser, D.~Gonzalez-Aguirre, F.~Sch\"ultje, J.~B. Sol, N.~Vahrenkamp, and
  T.~Asfour, ``Extracting whole-body affordances from multimodal exploration,''
  in \emph{Proceedings of the IEEE-RAS International Conference on Humanoid
  Robots}, 2014.

\bibitem{lemerle2011new}
D.~Mansour, A.~Micaelli, A.~Escande, and P.~Lemerle, ``A new optimization based
  approach for push recovery in case of multiple noncoplanar contacts,'' in
  \emph{Humanoid Robots (Humanoids), 2011 11th IEEE-RAS International
  Conference on}.\hskip 1em plus 0.5em minus 0.4em\relax IEEE, 2011, pp.
  331--338.

\bibitem{worgotter_simple_2013}
F.~W\"org\"otter, E.~E. Aksoy, N.~Kr\"uger, J.~Piater, A.~Ude, and
  M.~Tamosiunaite, ``A simple ontology of manipulation actions based on
  hand-object relations,'' \emph{Autonomous Mental Development, IEEE
  Transactions on}, vol.~5, no.~2, pp. 117--134, 2013.

\bibitem{KITDatabase}
``Kit whole-body human motion database,'' cited Feb. 2015,
  https://motion-database.humanoids.kit.edu/.

\bibitem{Mandery2015}
C.~Mandery, O.~Terlemez, M.~Do, N.~Vahrenkamp, and T.~Asfour, ``The {KIT}
  whole-body human motion database,'' in \emph{International Conference on
  Advanced Robotics (ICAR)}, 2015.

\bibitem{Terlemez2014_MMM}
O.~Terlemez, S.~Ulbrich, C.~Mandery, M.~Do, N.~Vahrenkamp, and T.~Asfour,
  ``{M}aster {M}otor {Map} ({MMM}) — framework and toolkit for capturing,
  representing, and reproducing human motion on humanoid robots,'' in
  \emph{Proceedings, IEEE-RAS International Conference on Humanoid Robotics
  (Humanoids)}, 2014, pp. 894--901.

\bibitem{ijspeert2002movement}
A.~J. Ijspeert, J.~Nakanishi, and S.~Schaal, ``Movement imitation with
  nonlinear dynamical systems in humanoid robots,'' in \emph{Robotics and
  Automation, 2002. Proceedings. ICRA'02. IEEE International Conference on},
  vol.~2.\hskip 1em plus 0.5em minus 0.4em\relax IEEE, 2002, pp. 1398--1403.

\bibitem{sentis_2010_compliant_Springer}
L.~Sentis, ``Compliant control of whole-body multi-contact behaviors in
  humanoid robots,'' in \emph{Motion Planning for Humanoid Robots}.\hskip 1em
  plus 0.5em minus 0.4em\relax Springer, 2010, p. 29–66.

\end{thebibliography}
%

\end{document}